\documentclass[letterpaper, 10 pt, conference]{ieeeconf}
\IEEEoverridecommandlockouts
\usepackage[backend=biber,
            hyperref=true,
            url=false,
            isbn=false,
            doi=false,
            backref=false,
            style=ieee,
            citestyle=numeric-comp,
            sorting=nyt,
            block=none]{biblatex}
\usepackage{multicol}
\usepackage{multirow}
\usepackage{xspace}
\usepackage{xcolor}
\usepackage{graphicx}
\usepackage{caption}
\usepackage{subcaption}
\usepackage{adjustbox}
\usepackage{xfrac}
\usepackage{listings}
\usepackage{float}
\usepackage{booktabs}
\usepackage{bm}
\usepackage{amsfonts}
\usepackage{tcolorbox}
\usepackage{hyperref}
\usepackage[moderate,tracking=normal]{savetrees}

\newcommand{\q}[1]{$\mathbf{\mathcal{Q}#1}$}

\title{\LARGE \bf
Chain-of-Modality: Learning Manipulation Programs from\\Multimodal Human Videos with Vision-Language-Models
}

\author{Chen Wang$^{1\dagger}$, Fei Xia$^{1}$, Wenhao Yu$^{1}$, Tingnan Zhang$^{1}$, Ruohan Zhang$^{2}$,\\C. Karen Liu$^{2}$, Li Fei-Fei$^{2}$, Jie Tan$^{1}$, Jacky Liang$^{1}$
\thanks{$^{1}$Google DeepMind $^{2}$Stanford University}
\thanks{$^{\dagger}$Work done while interning at Google DeepMind}
}

\begin{document}

\maketitle
\thispagestyle{empty}
\pagestyle{empty}

\begin{abstract}
Learning to perform manipulation tasks from human videos is a promising approach for teaching robots. However, many manipulation tasks require changing control parameters during task execution, such as force, which visual data alone cannot capture. In this work, we leverage sensing devices such as armbands that measure human muscle activities and microphones that record sound, to capture the details in the human manipulation process, and enable robots to extract task plans and control parameters to perform the same task. To achieve this, we introduce Chain-of-Modality (CoM), a prompting strategy that enables Vision Language Models to reason about multimodal human demonstration data --- videos coupled with muscle or audio signals. By progressively integrating information from each modality, CoM refines a task plan and generates detailed control parameters, enabling robots to perform manipulation tasks based on a single multimodal human video prompt. Our experiments show that CoM delivers a threefold improvement in accuracy for extracting task plans and control parameters compared to baselines, with strong generalization to new task setups and objects in real-world robot experiments. Videos and code are available at \href{https://chain-of-modality.github.io}{chain-of-modality.github.io}
\end{abstract}

\vspace{5.0pt}

\section{INTRODUCTION}

Can robots learn to perform physically challenging manipulation tasks (e.g., twisting to open water bottles or drumming) by watching only one human hand video demonstration? One way to enable this capability is through recognizing human task plans from video and then translating them into executable robot skills. 
While recent advances in video understanding show promising results in action recognition~\cite{kong2022human}, many manipulation skills require precise specification of the control parameters, which cannot easily be inferred from pure visual information, e.g. grasp \textit{lightly} to rotate a key in hand, push \textit{harder} to insert a plug, hit a drum \textit{gently} to produce a soft sound. This limitation restricts robots' ability to perform diverse manipulation tasks by only watching human videos.

A core challenge here is that extracting task plans from human video data is difficult since vision-only data lacks necessary details in recognizing these plans. A key observation is that many details in human task plans, such as control parameters like force and speed, are better captured through additional signals such as human muscle activities and object interaction sound. For example, when inserting a power plug, humans would first \texttt{Grasp} the plug with \textit{low force} to adjust its orientation in hand, then hold it \textit{firmly} while \texttt{Insert} it into the socket. In this work, we leverage sensing devices such as modern armbands equipped with muscle sensors and sports cameras with microphones to collect multimodal demonstration videos that include images, muscle activity, and object interaction sounds. They provide additional information about \textit{when} and \textit{how} humans exert physical effort during manipulation. However, effectively utilizing these signals requires new methods for reasoning from multimodal human demonstration videos.

Vision-language models (VLMs) are capable of solving a broad range of practical problems, from visual reasoning to signal processing~\cite{yang2023dawn} and even code generation for controlling robots~\cite{liang2023code}. Recent advancements in long-context input have further enabled VLMs to take videos and long-sequence numerical signals as inputs~\cite{chiang2024mobility}. This makes us ask: Can VLMs serve as general reasoning models to infer human task plans from multimodal demonstration videos? Most VLM applications still only take one modality as input. To tackle this challenge, we introduce Chain-of-Modality (CoM), a framework that prompts VLMs to analyze each modality one after another, progressively refining the answer by incorporating new information from each modality.

\begin{figure}[t]
  \centering
  \includegraphics[width=1.0\columnwidth]{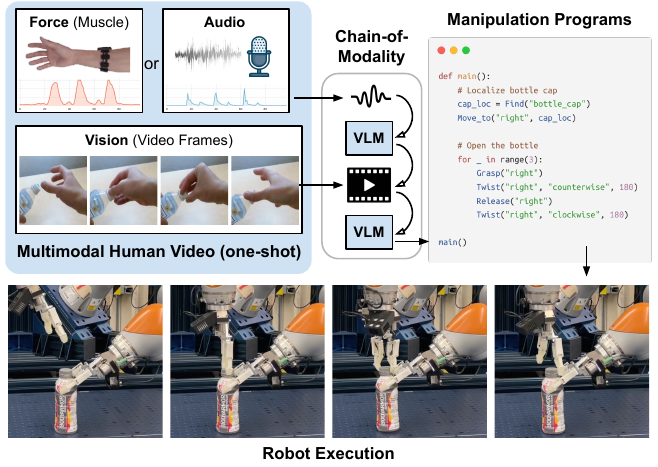}
  \caption{We introduce Chain-of-Modality (CoM), a prompting strategy that enables VLMs to recognize human task plans from a single multimodal video with force or audio information, and generate corresponding robot control code to reproduce the task.}
  \label{fig:pull}
\end{figure}

\begin{figure*}[t]
  \centering
  \includegraphics[width=2.0\columnwidth]{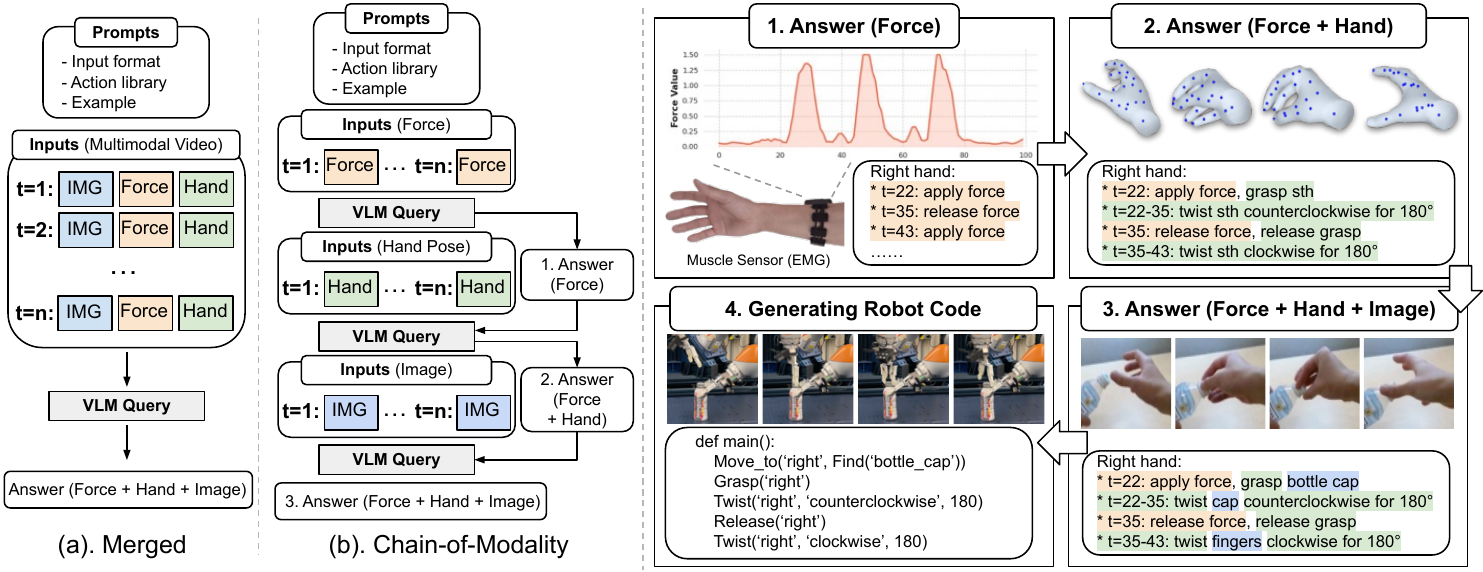}
  \caption{Overview of Chain-of-Modality (using force as an example). \textbf{(a) Baseline method - Merged}: Merges multimodal information (vision, force, and hand pose) into a single input batch and queries the VLM to directly generate the final answer. \textbf{(b) Chain-of-Modality (CoM)}: Analyzes each modality step-by-step, refining the analysis to produce the final answer. \textbf{Example}: First, the VLM uses force data to determine when force is applied. Then, with hand pose information, it infers that the human is grasping and twisting. Next, with image data, the VLM identifies the action as twisting a bottle cap. Finally, VLM transform the CoM analysis into a robot-executable Python program to reproduce the task.}
  \label{fig:method}
\end{figure*}

CoM enables VLMs to extract task plans and control parameters by analyzing a single multimodal human video. With CoM, the inclusion of additional modalities helps VLMs better segment subtasks. For instance, when a human is opening a bottle, three peak signals in the force data indicate three \texttt{Twist} motions. CoM enables the VLMs to leverage this information to first segment the entire task into a coarse task skeleton and progressively filling in more details by incorporating other modalities. Additionally, force information acquired from additional modalities allows VLMs to generate more accurate control parameters for skills like \texttt{Grasp} and \texttt{Hit} with different level of force. Empirically, we found that CoM has achieved a $60\%$ accuracy in extracting \emph{exact} task plans and control parameters from human video. Methods that rely solely on vision-only data have zero accuracy, and na\"ive methods that directly query the model with all modalities in a single batch achieve an average accuracy of $17\%$.

How do we turn these task plans into robot actions? Prior work \cite{liang2023code} has demonstrated that foundation models can generate robot-executable API calls based on language instructions. Our setting is different in the sense that we ask foundation models to generate API calls based on a multimodal demonstration video. These API calls provide benefits of cross-embodiment generalization since the code API can abstract away robot embodiment, allowing smooth deployment across different robots. In addition, code API based on advanced vision models further enable the robot to generalize to novel objects and unseen object configurations.

Our main contributions are as follows:
\begin{itemize}
    \item \textbf{Chain-of-Modality (CoM)}: A prompting strategy that enables VLMs to reason from multimodal human video demonstration data by progressively incorporating vision and force information.
    \item \textbf{One-shot manipulation program generation}: A pipeline for generating robot control programs from a single multimodal human demonstration video, integrating force information (obtained through muscle or audio signals) to produce fine-grained control parameters of different skills.
    \item \textbf{Generality}: We show the benefit of CoM is consistent across two advanced VLM models, and that our method allows the VLMs to learn from a single human video to write robot code that works on different real-world robot platforms with generalization capabilities.
\end{itemize}

\section{RELATED WORK}

\textbf{Human Activity Understanding from Video.} Understanding human activities in video has been a long-standing focus in computer vision~\cite{caba2015activitynet, goyal2017something, Damen2018EPICKITCHENS, li2019hake, grauman2022ego4d}. Early works primarily aimed to capture the high-level semantic meaning of videos through classification~\cite{xu2017r, wang2018temporal}. In efforts to extract more detailed information, later studies began to focus on deriving task plans from videos~\cite{zhang2022actionformer, huang2019neural, chang2020procedure}. However, these methods tend to be limited by their reliance on specific training datasets, making it challenging to generalize to unseen action categories. In recent years, the development of large vision-language models has led to impressive results in prompting VLMs to understand human activities from videos~\cite{ju2022prompting}. Unlike prior works, our work focuses on reasoning multimodal human videos with force or audio information, providing essential information for downstream fine-grained robot manipulation tasks.

\textbf{Foundation Models for Robotics and Control.} In recent years, foundation models have demonstrated significant progress in robotics, spanning high-level reasoning to low-level control~\cite{firoozi2023foundation, hu2023toward}. Earlier works primarily focused on language-conditioned robotic reasoning and planning, where tasks were defined using natural language~\cite{huang2022language, zeng2022socratic, ahn2022can, huang2022inner, raman2022planning, silver2024generalized, liu2023llm+, lin2023text2motion, wang2023describe, chen2024spatialvlm}. However, some manipulation tasks—especially those involving spatial ambiguity or requiring fine-grained control—are difficult to specify using language alone. Recent advances in vision-language models (VLMs) have introduced more expressive task specifications, such as visual annotations~\cite{nasiriany2024pivot, gu2023rttrajectory, sundaresan2024rt, huang2024rekep}. Our work, on the other hand, uses single-shot multimodal video as task specifications, enabling robots to extract task plans and control parameters from human demonstrations. To apply foundation models to robot control, several promising approaches have emerged, including subgoal selection for goal-conditioned policies~\cite{shah2023lm, cui2022can, shridhar2022cliport}, reward or constraint generation for trajectory optimization~\cite{yu2023language, liang2024learning, huang2024rekep}, and code generation based on perceptual and control primitives~\cite{liang2023code, singh2023progprompt}. Unlike these approaches, which are conditioned on language inputs, we demonstrate how VLMs can directly reason from \textit{one-shot} human video inputs to generate low-level manipulation programs, providing an alternative way to prompt robot to perform new tasks with rich visual hints.

\textbf{Learning Manipulation from Human Video.}
A plethora of recent research has explored leveraging human video data to teach robot manipulation skills~\cite{xiong2021learning,das2021model,zakka2022xirl,shao2021concept2robot,sharma2019third,smith2019avid,schmeckpeper2020learning,edwards2019perceptual,schmeckpeper2020reinforcement, wen2022you, bahl2022human, kumar2023graph, videodex, wang2023mimicplay, xu2023xskill, bharadhwaj2024towards, jain2024vid2robot, papagiannis2024r+}. These works focus on extracting different information from human video, such as object affordance~\cite{koppula2015anticipating, bahl2023affordances}, motion trajectories~\cite{wang2023mimicplay, wen2023anypoint, zhu2024vision}, task dynamics~\cite{rhinehart2017first, liu2018imitation}, and reward representations~\cite{sermanet2016unsupervised, sermanet2018time, chen2021learning, ma2022vip}. Works like~\cite{wang2023mimicplay, jain2024vid2robot, zhu2024vision} train manipulation policies that are conditioned on human or robot videos, instead of language instructions. Despite their effectiveness, because such methods only learn from videos (sequence of images), they cannot infer important details, such as how much force to apply, required for many manipulation tasks. In this work, we focus on developing methods that can leverage multiple sensing modalities, including images, forces, and sounds, to better understand the subtle details in human demonstrations that are not readily visible, and to enable robots to better perform such tasks.

\begin{figure*}[t]
  \centering
  \includegraphics[width=2.0\columnwidth]{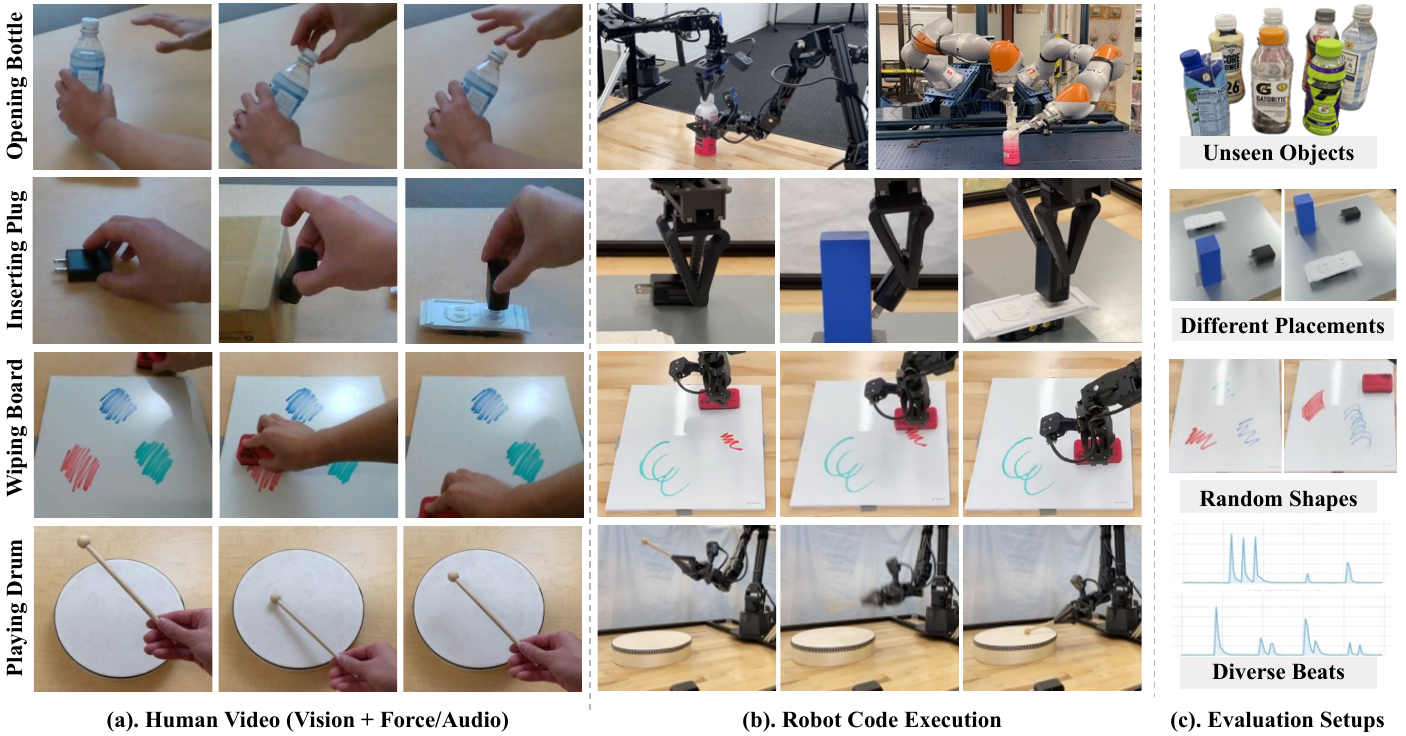}
  \caption{Overview of Experiment Tasks. \textbf{(a) Multimodal Human Video Input}: Our framework processes a single-shot human video with force or audio data, using Chain-of-Modality to extract the task plan and control parameters, then generates a robot control program. \textbf{(b) Robot Code Execution}: The robot executes the program to replicate the task observed in the video. \textbf{(c) Evaluation Setups}: We evaluate the performance of generated program in various experimental setups.}
  \label{fig:task}
\end{figure*}

\section{Learning from Multimodal Human Videos}

We introduce our system design, which takes a single multimodal human demonstration video as input, and generates a robot executable code to perform the manipulation task demonstrated in the video. The system has three main components: (1) collecting multimodal human videos; (2) Chain-of-Modality for understanding multimodal human videos; and (3) generating code and controlling the robot. For each component, we first discuss the motivation, followed by examples.

\subsection{Multimodal Human Demonstration Video}

Videos often struggle to capture important details of fine-grained details of a human performing a manipulation task, especially those involving force application. For example, when inserting a power plug (Fig.~\ref{fig:task}), we first apply \textit{light} force to adjust its orientation, then \textit{increase} force for insertion. These varying force levels are critical but hard to observe from video alone, highlighting the need for multimodal data that goes beyond visual information.

To address these challenges, we consider multimodal human video consists of, at each timestep, an RGB image, human muscle signal or object interaction sound, and hand pose (Fig.~\ref{fig:pull}). They collectively provide a more comprehensive view of human task plans. Human muscle signals captured by an armband with muscle sensors (EMG) or object interaction sound captured by a microphone can provide necessary force information, which indicates the timing and the amount of force human applied during the entire task. Moreover, to provide more detailed information on human hand motions, we use a vision-based method~\cite{pavlakos2024reconstructing} to estimate hand pose and treat the pixel locations of the fingertips as another input modality.

\subsection{Chain-of-Modality}

Next, we use Vision Language Models (VLMs) to analyze the rich information provided in such multimodal human videos to extract task plan descriptions. The VLM needs to process signals from all these modalities: recognizing the human actions in the correct temporal order and determining the control parameters of each action (e.g., name of the target object, direction of the motion). One way to use a VLM for this purpose is to directly query the model with all modalities interleaved together in a sequence (Fig.~\ref{fig:method}(a)). However, we found that state-of-the-art VLMs (e.g. Gemini 1.5 Pro~\cite{reid2024gemini}, GPT-4o~\cite{achiam2023gpt}) often struggle to correlate information between modalities, leading to issues like neglecting certain inputs or attempting to extract information from the wrong modality. To improve VLMs' performance in understanding multimodal human videos, we propose Chain-of-Modality (CoM, Fig.~\ref{fig:method}(b)), a prompting strategy that queries the VLM to analyze each modality sequentially, extracting key information and progressively aggregating results to produce the final answer.

\textbf{Prompting Chain-of-Modality.}
The CoM prompt consists of three parts: (1) descriptions of each modality and their input data format, (2) description of the available action set along with an explanation of action parameters, and (3) one example of video-to-analysis pair introducing how each modality should be analyzed to produce a sequence of recognized actions with parameters.

\textbf{Examples of Chain-of-Modality.}
Fig.~\ref{fig:method} illustrates an example of using CoM to analyze a multimodal human video. In this video, a person holds the bottle with the left hand and twists the cap with the right hand. CoM sequentially analyzes each input modality and refines the answer based on the prior analysis. In Fig.~\ref{fig:method}, we highlight the new information contributed by each modality with a separate color.
In the first stage, the VLM analyzes force or auditory signals and finds out \textit{when} the person applies and releases force. Then it deduces the number of times the person applied force. However, without hand and image information, it is unclear what the person is doing exactly.
In the second stage, the VLM incorporates hand pose information. It now recognizes that the person is \emph{grasping} and \emph{twisting} during force application. The finger position also indicates a \emph{counterclockwise twist} of about 180 degrees while \emph{holding} the bottle and a \emph{clockwise finger rotation} during release. Still, without image data, the objects that appeared in the task remain unknown.
In the third stage, the VLM integrates image data. It identifies that the left hand is holding \emph{the bottle} and the right hand is twisting \emph{the cap}. With this information, the VLM generates action functions, specifying detailed action parameters at each time step.
Note that, none of the tasks or objects appear in the example prompts. The example prompts are only used to demonstrate the output format of the analysis and the library of available skills.

\subsection{Writing Robot Code}

Based on the human video analysis obtained above, the final step is to transform the action sequence into robot-executable code with low-level API calls. We use the same VLMs to perform this code generation~\cite{liang2023code} to create manipulation programs to accomplish the task. Prompts for code generation include the video analysis as well as descriptions of robot APIs and the required output format e.g.,
\lstset{language=Python}
\begin{lstlisting}
from skills import Grasp, Release, Twist, Find, Move_to
# Based on video analysis and APIs, generate python code:
\end{lstlisting}

\textbf{Examples of Generating High-Level Task Plans.} Below is an example of the generated program for the aforementioned bottle-opening task:
\lstset{language=Python}
\begin{lstlisting}
Move_to('left', Find('bottle'))
Grasp('left')
Move_to('right', Find('bottle_cap'))
for _ in range(3):
    Grasp('right')
    Twist('right', 'counterclockwise', 180)
    Release('right')
    Twist('right', 'clockwise', 180)
\end{lstlisting}
VLM incorporates the video analysis from CoM and generates a detailed task plan for opening a bottle, which includes using the right gripper to \texttt{Twist} counterclockwise while holding the cap and \texttt{Twist} clockwise without holding the cap. It also generates a for-loop script to specify the periodic twisting motion. 

\textbf{Examples of Generating Control Parameters.} Besides generating task plans, in contact-rich tasks such as inserting a plug into a power strip, VLM can also generate control parameters to specify the use of force:
\lstset{language=Python}
\begin{lstlisting}
from skills import Grasp, Push_towards, Insert
Grasp('right', 'plug', 100) # force range from [0, 100]
Move_to('right', 'box', 20) # rotate plug in-hand
Insert('right', 'power_strip', 100)
\end{lstlisting}
Leveraging the force information from the multimodal human video, VLM specifies the amount of force to apply during different task stages, allowing the use of \texttt{Move\_to} to re-orient the plug in hand by pushing towards the wall (force = 20) and to hold the plug \textit{firmly} while \texttt{Insert} it into the power strip (force = 100). 

\begin{figure*}[t]
  \centering
  \includegraphics[width=2.0\columnwidth]{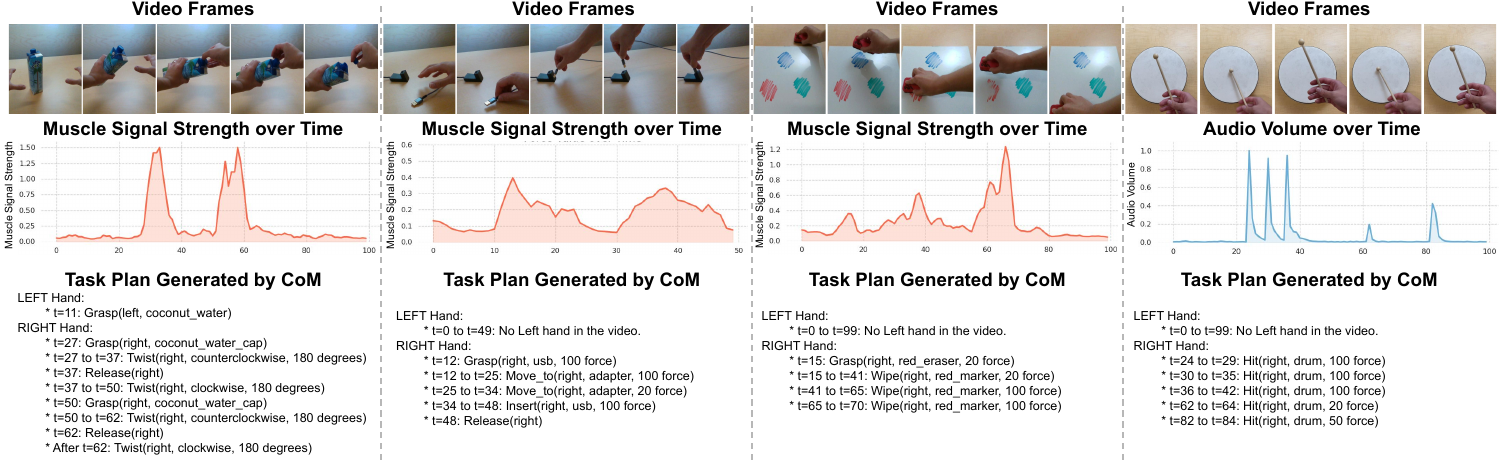}
  \caption{Qualitative results for Chain-of-Modality. We showcase task plans generated by CoM for four evaluation videos. CoM successfully segments the videos into subtasks, specifying the skills, force, and target objects at each stage.}
  \label{fig:results}
\end{figure*}

\begin{figure*}[t]
  \centering
  \includegraphics[width=2.0\columnwidth]{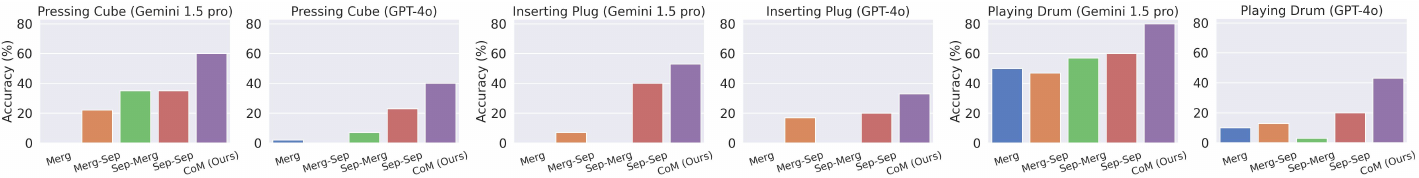}
  \caption{Quantitative results for Chain-of-Modality. We compare CoM with baselines across three tasks using both Gemini and GPT.}
  \label{fig:results_com}
\end{figure*}

\subsection{Implementation Details}

\textbf{Data Collection.} The muscle signal (EMG) contains eight channels of data sampled at 200Hz. Since the camera records at 60Hz, we downsample the muscle signals to match the camera sampling rate and use the maximum value across the eight channels as the force signal at each time step. Similar to audio signals, we compute the loudness of the sound at each time step as the input audio value. For the hand pose estimation, we use HaMeR~\cite{pavlakos2024reconstructing} to localize the pixel locations of the fingertips. More details of the signal processing steps can be found in the Appendix.

\textbf{Robot Execution.} The robot API calls consist of predefined control functions that ``ground'' the generated programs in a real-world robot system. These APIs benefit significantly from recent advances in perception models. For instance, in our experiments, all object localization is performed by querying Gemini 1.5 Pro with an RGB-D image and the name of the target object (as specified in the generated programs), which directly generates a 2D bounding box around the target object on the RGB image. We then use the depth information and camera parameters to create 3D point clouds of the entities within the detected bounding box and use the average 3D location to represent the object's position. These open-vocabulary APIs simplify the connection between the generated programs and the robot perception system, directly enhancing the capability of the code-based robot policy. More details of the robot APIs can be found in the Appendix.

\section{Experiments}

Our experiments aim to answer the following questions:
\begin{enumerate}
    \item[\q{1}:] Does Chain-of-Modality improve VLMs' understanding of multimodal human videos?
    \item[\q{2}:] Does force information help VLMs in reasoning about human task plans?
    \item[\q{3}:] Does hand pose aid fine-grained manipulation?
    \item[\q{4}:] Can VLMs with CoM extract control parameters from multimodal human videos?
    \item[\q{5}:] How does the generated program perform on real robots?
\end{enumerate}

\subsection{Experiment Setups}

\textbf{Baselines.} We categorize the baselines into two groups: baselines with different input modalities and baselines with different VLM reasoning procedures.
For methods with different input modalities, the baselines include: \textit{(1) Image-only} – uses only video image inputs, \textit{(2) w.o. img} – uses force and hand pose inputs without image inputs, \textit{(3) w.o. force} – excludes force input, \textit{(4) w.o. hand} – excludes hand pose input, and \textit{(5) All} – utilizes all three input modalities: force, hand pose, and image.
For methods with different VLM reasoning procedures, the baselines include: \textit{(1) Merg} – merges all modality inputs and directly generates the final answer, \textit{(2) Merg-Sep} – merges all inputs but generates separate answers for each modality, \textit{(3) Sep-Merg} – separate each modality inputs, then directly generates one final answer, \textit{(4) Sep-Sep} – separates each modality input and generates separate answers for each, followed by a final answer, and \textit{(5) Ours} – processes each modality sequentially, analyzing each modality based on the analysis of the previous one until the final answer.

\textbf{Tasks.} We design two categories of tasks: multimodal video analysis and real-world robot evaluation with generated programs.
In multimodal video analysis, we test all methods on four tasks: \textit{Pressing Cube}, \textit{Inserting Plug}, \textit{Playing Drum}, and \textit{Opening Bottle}. These tasks feature long task horizons, force sensitivity, and bi-manual manipulation. Each task consists of 10 testing videos with varying objects and camera viewpoints. For each baseline, we query VLM three times per video and report the average success rate in generating the correct human task plans observed in the video, along with the similarity score, which is calculated by finding the longest common string between the output and the ground truth. For all four tasks, we test with two advanced VLM models: Gemini 1.5 Pro \cite{reid2024gemini} and GPT-4o \cite{achiam2023gpt}.
In real-world robot evaluation, we test four tasks (Fig.~\ref{fig:task}): \textit{Opening Bottle}, \textit{Inserting Plug}, \textit{Wiping Board} and \textit{Playing Drum}. Each method is tested with 20 trials, and we report the average success rate.

For testing the generalization capability, as Fig.~\ref{fig:task}(c) illustrated, we test \textit{Opening Bottle} with 7 types of bottle (6 unseen); we also randomly placed the plug, power strip, and box in \textit{Inserting Plug task}. In \textit{Wiping Board}, we draw marker with different shapes with varying positions on the board. In \textit{Playing Drum} task, we test different drumming beats. For testing the cross-embodiment deployment of our method, we test the \textit{Opening Bottle} task on two robot platforms: the bi-manual ViperX and the bi-manual KUKA.

\textbf{Prompts.} For video analysis tasks, the prompts we use include an explanation of the input data format and one example video. The input format explains the hand pose data, which consists of 2D pixel locations of the thumb and middle fingertips. The muscle or audio signals is normalized to a single float value.
The example video involves a human interacting with task-irrelevant objects in random manner. For instance, both the \textit{Pressing Cube} and \textit{Opening Bottle} tasks use the same example video, where the human is pressing and rotating an apple and a can on the table. This video demonstrates only the key features of primitive skills, \emph{which does not include the testing objects or task plans}.
We also provide expected output for the example video, helping the VLM generate responses in the same format, which facilitates benchmarking the results. In each evaluation task, all baseline methods use the same example video.
After prompting, we provide the testing video without task description and directly let VLM generate its analysis of the input multimodal video.
For real-world robot evaluation, the prompts include the video analysis previously generated by the our CoM pipeline, along with the definitions of robot API calls in Python format. We then ask the VLM to generate the main function to reproduce human task plans from the video analysis. After the VLM generates the program, we use \textit{exec()} to execute the generated code on the real-world robot.

\begin{table}[t]
\centering
\resizebox{\linewidth}{!}{
\footnotesize
\setlength{\tabcolsep}{0.8mm}{
\begin{tabular}{cl|cccc}
\toprule
\multicolumn{1}{l}{}                                                      &            & \begin{tabular}[c]{@{}c@{}}Pressing\\Cube\end{tabular} & \begin{tabular}[c]{@{}c@{}}Opening\\Bottle\end{tabular} & \begin{tabular}[c]{@{}c@{}}Inserting\\Plug\end{tabular} & \begin{tabular}[c]{@{}c@{}}Playing\\Drum\end{tabular} \\
\midrule
\multirow{5}{*}{\begin{tabular}[c]{@{}c@{}}Gemini 1.5\\ Pro\end{tabular}} & Image-only & 0.00/0.68          & 0.00/0.32           & 0.00/0.80    &    0.00/0.31    \\
                                                                          & w.o. img   & 0.00/0.00          & 0.00/0.45           & 0.00/0.00    &    0.00/0.00   \\
                                                                          & w.o. force & 0.00/0.68          & 0.00/0.64           & 0.00/0.72    &   0.00/0.03    \\
                                                                          & w.o. hand  & \textbf{0.70}/\textbf{0.96}          & 0.00/0.49           & 0.47/\textbf{0.96}   &   0.57/0.90    \\
                                                                          & All       & 0.67/0.92          & \textbf{0.37}/\textbf{0.75}           & \textbf{0.53}/0.93     &  \textbf{0.80}/\textbf{0.96}    \\
\midrule
\multirow{5}{*}{GPT-4o}                                                   & Image-only & 0.00/0.48          & 0.00/0.43           & 0.00/0.77     &   0.00/0.37   \\
                                                                          & w.o. img   & 0.00/0.00          & 0.00/0.53           & 0.00/0.00     &   0.00/0.00   \\
                                                                          & w.o. force & 0.00/0.13          & 0.00/0.20           & 0.00/0.55     &   0.00/0.22   \\
                                                                          & w.o. hand  & \textbf{0.43}/0.76          & 0.00/\textbf{0.54}           & 0.30/\textbf{0.81}     &  0.34/0.84    \\
                                                                          & All       & 0.40/\textbf{0.84}          & 0.00/0.46           & \textbf{0.33}/0.79   &  \textbf{0.43}/\textbf{0.86}\\      
\bottomrule
\end{tabular}
}}
\caption{Multimodal eval (Accuracy / Similarity Score).}
\label{tab:multimodal}
\vspace{-10.0pt}
\end{table}

\begin{table}[]
\resizebox{\linewidth}{!}{
\footnotesize
\setlength{\tabcolsep}{0.8mm}{
\begin{tabular}{l|cccccc}
\toprule
           & \begin{tabular}[c]{@{}c@{}}Opening Bottle\\(ViperX)\end{tabular} & \begin{tabular}[c]{@{}c@{}}Opening Bottle\\(KUKA)\end{tabular} & \begin{tabular}[c]{@{}c@{}}Insert\\Plug\end{tabular} & \begin{tabular}[c]{@{}c@{}}Wiping Board \\ (red marker)\end{tabular} & \begin{tabular}[c]{@{}c@{}}Wiping Board \\ (blue marker)\end{tabular} & \begin{tabular}[c]{@{}c@{}}Playing\\Drum\end{tabular} \\ \hline
Ours       &  12/20         & 15/20            &     15/20         &       16/20     &       14/20     &   16/20     \\ 
Oracle               &  16/20         & 20/20            &     18/20         &       20/20     &       16/20     &   20/20     \\
\bottomrule
\end{tabular}
}}
\caption{On-robot evaluation results. All results are obtained in the generalization settings introduced in the \textbf{Task} section.}
\label{tab:robot}
\vspace{-10.0pt}
\end{table}

\subsection{Results}

\textbf{Chain-of-Modality helps in understanding multimodal human videos.} In Fig.~\ref{fig:results_com}, we compare baseline methods and Chain-of-Modality across three tasks using both Gemini 1.5 Pro and GPT-4o. We first observe that processing and analyzing each modality separately consistently outperforms other baselines that either merge the modality inputs or generate a single merged answer. Both \textit{Sep-Sep} and \textit{CoM} achieve the best performance in all tasks with different VLMs. \textit{CoM} further outperforms \textit{Sep-Sep} by more than 19\% with Gemini 1.5 Pro and 17\% with GPT-4o. This result indicates that although advanced VLMs have long-context inputs, directly reasoning over long-context inputs remains challenging. \textit{CoM}'s per-modality analysis and progressively fulfilling the final answer is more suitable for current VLMs to reason from multimodal human videos.

\textbf{Force information benefits learning from human videos.} In Tab.~\ref{tab:multimodal}, we compare methods with different input modalities. We observe that images play a crucial role in recognizing task objects, as \textit{w.o. img} baseline achieve no success. For tasks like \textit{Opening bottle}, the main skill \textit{Twist} doesn't rely on object names, yielding a decent similarity score, but in general, images are essential. Force information greatly enhances the understanding of human task plans. Methods with force inputs (\textit{w.o. hand} and \textit{All}) significantly outperform those without. This indicates that extracting key information for manipulation purely from images is still challenging. Force inputs help VLMs better segment the video into different stages, leading to an average of 42\% improvement in the similarity score between the extracted task plan and the ground truth (\textit{All} v.s. \textit{w.o. force}). Without force information, the baselines achieve no success in predicting the correct task plans and skill parameters from human videos. 

\textbf{Hand pose helps understand fine-grained manipulation.} In the \textit{Opening Bottle} task, only the method with all modalities as input (\textit{All}) achieves a non-zero success rate. This task requires extracting fine-grained information about fingertip rotation direction, along with multiple instances of grasping and releasing the bottle cap (as illustrated in the qualitative results, Fig.~\ref{fig:results_com}, first column). Hand pose plays a crucial role in this task, indicating that current VLM models are not yet proficient at extracting human hand motions. The hand pose estimated by specialized vision models~\cite{pavlakos2024reconstructing} provides significant assistance.

\textbf{CoM is capable of extracting control parameters from multimodal human videos.} In the qualitative results shown in Fig.~\ref{fig:results_com}, CoM successfully (1) identifies task plans and target objects for single-arm or bi-manual tasks from multimodal human videos and (2) extracts detailed control parameters, including motion direction, force intensity, and timing of each action. This information paves the way for generating robot executable programs to perform the same task.

\textbf{VLMs can generate manipulation programs based on CoM analysis.} Although CoM can generate task plans from a single human video, these plans still lack detailed robot API calls, such as \textit{Find} the target object and \textit{Move\_to} the object's location before \textit{Grasp}. VLMs can fill in these missing sub-steps and generate final robot-executable Python API calls. In Tab.~\ref{tab:robot} and our video submission, we demonstrate how the final generated program controls the robot to perform the task with an average success rate of 73\% across all tasks, showing strong generalization to novel objects (e.g., opening 8 unseen bottles, wiping a board, and inserting a power plug with randomized object placements). This video demonstrates only the key features of primitive skills, \emph{which does not include the testing objects or task plans}. The \textit{Oracle} results (92\%) showcases the upper bound of the robot system with manually-created robot code for each task. Notably, the \textit{Find} API call we use is implemented by directly querying Gemini 1.5 pro, which generates 2D bounding boxes from free-form language inputs and further processes them into 3D locations using RGB-D images. This highlights the potential of leveraging VLMs for straightforward implementation of low-level robot API calls. In the \textit{Opening Bottle} task, we tested our framework on two types of bi-manual robots (ViperX and KUKA), demonstrating the potential for cross-embodiment deployment of our method.

\section{Conclusion and Limitations}

We present Chain-of-Modality (CoM), a prompting strategy that enables Vision-Language Models (VLMs) to understand multimodal human video demonstration data by combining video with force or audio inputs. By progressively refining task plans and control parameters through each modality, CoM enhances robots' ability to perform one-shot imitation from human videos in fine-grained manipulation tasks. Our experiments show that CoM significantly improves the accuracy of task plan recognition and control parameter extraction compared to baselines, with strong generalization to new task setups and objects in real-world robot experiments.

The limitations of this work include: (1) The audio modality we use focuses only on the volume of impact sounds, without fully capturing other aspects such as frequency and pitch. Future work could explore the broader use of audio in foundation models for multimodal human video reasoning. (2) In this work, we focus on extracting task plans and control parameters from human videos and executing them on the robot in an open-loop fashion. In future work, we plan to explore generating closed-loop control programs that can adapt to unexpected situations.

\printbibliography

\end{document}